\documentclass[1p, review]{elsarticle}

\usepackage[english]{babel}
\usepackage[utf8]{inputenc}
\usepackage{amsmath}
\usepackage{amssymb}
\usepackage{amsfonts}
\usepackage{graphicx}
\usepackage{epstopdf}
\usepackage{subcaption}
\usepackage[justification=centering]{caption}
\usepackage{multirow}
\usepackage{booktabs}
\usepackage{lineno}
\usepackage{float}
\usepackage{transparent}

\usepackage{algorithm}
\usepackage{algpseudocode}
\algnewcommand\algorithmicinput{\textbf{Input:}}
\algnewcommand\Input{\item[\algorithmicinput]}

\usepackage{hyperref}
\hypersetup{
    colorlinks,
    citecolor=blue,
    filecolor=blue,
    linkcolor=blue,
    urlcolor=blue
}

\newdefinition{rmk}{Remark}

\usepackage[procnames]{listings}
\usepackage{color}
\definecolor{keywords}{RGB}{255,0,90}
\definecolor{comments}{RGB}{0,0,113}
\definecolor{red}{RGB}{160,0,0}
\definecolor{green}{RGB}{0,150,0}
 
\begin{document}
\journal{Journal of Computer Vision and Image Understanding}

\begin{frontmatter}

\title{Randomized Low-Rank Dynamic Mode Decomposition for Motion Detection}

\author[rvt]{N. Benjamin~Erichson\corref{cor1}}
\ead{nbe@st-andrews.ac.uk}
\author[rvt]{Carl~Donovan}
\address[rvt]{School of Mathematics and Statistics, Univ. of St Andrews, UK}
\cortext[cor1]{Corresponding author}

\begin{abstract}
This paper introduces a fast algorithm for randomized computation of a low-rank Dynamic Mode Decomposition (DMD) of a matrix.  Here we consider this matrix to represent the development of a spatial grid through time e.g. data from a static video source. DMD was originally introduced in the fluid mechanics community, but is also suitable for motion detection in video streams and its use for background subtraction has received little previous investigation. In this study we present a comprehensive evaluation of background subtraction, using the randomized DMD and compare the results with leading robust principal component analysis algorithms. The results are convincing and show the random DMD is an efficient and powerful approach for background modeling, allowing processing of high resolution videos in real-time. Supplementary materials include implementations of the algorithms in \textit{Python}.
\end{abstract}

\begin{keyword}
\texttt{dynamic mode decomposition; robust principal component analysis; randomized singular value decomposition; motion detection; background subtraction; video surveillance;}

\end{keyword}

\end{frontmatter}


\section{Introduction}
\begin{figure}[htp]
	\centering
	\DeclareGraphicsExtensions{.eps}
	\includegraphics[width=0.6\textwidth]{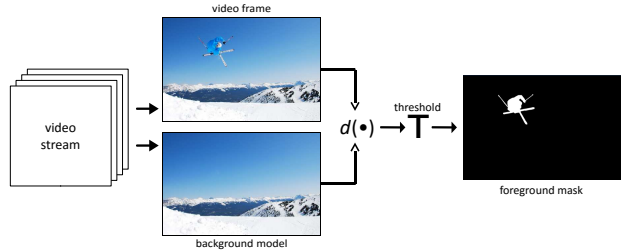}
	\caption{Illustration of background subtraction}
	\label{fig:backsub}
\end{figure}

The demand for video processing is rapidly increasing, driven by greater numbers of sensors with greater resolution, new types of sensors, new collection methods and an ever wider range of applications. For example, video surveillance, vehicle automation or wild-life monitoring, with data gathered in visual/infra-red spectra or SONAR, from multiple sensors being fixed or vehicle/drone-mounted etc. The overall result is an explosion in the quantity of high dimensional sensor data. Motion detection is often the fundamental building block for more complex video processing and computer vision applications, e.g. object tracking or human behavior analysis. In practice, there are many different types of sensors giving data suitable for object extraction, however we focus here on video data provided by static optical cameras, noting the findings generalize to other data types. In this case, the change in position of an object relative to its surrounding environment can be detected by intensity changes over time in a sequence of video frames. The challenge therefore is to separate intensity changes corresponding to moving objects from those generated by background noise i.e. dynamic and complex backgrounds. 
From a statistical point of view this can be formulated as a density estimation problem, aiming to find a suitable model describing the background. Moving objects can then be identified by differences from the reconstructed background from the video frames, via some thresholding, as illustrated in Figure~\ref{fig:backsub}.
In practice, the problem of finding a suitable model is difficult and often ill-posed due to the many challenges arising in real videos, e.g., dynamic backgrounds, camouflage effects, camera jitter or noisy images, to name only a few. One framework for tackling these challenges is provided by subspace learning techniques. Recently, robust principal component analysis (RPCA) has been very successful in separating video frames into background and foreground components \cite{bouwmans2014robust}. However, RPCA comes with relatively high computational costs and it is of limited utility for real-time analysis of high resolution video. Hence, in light of increasing sensor resolutions there is a need for algorithms to be more rapid, perhaps by approximating existing techniques. 

A competitive alternative is Dynamic Mode Decomposition (DMD) --- a data-driven method allowing decomposition of a matrix representing both time and space \cite{Kutz2013}. Due to the unique properties of videos (equally spaced time with high temporal correlation), DMD is well suited for motion detection, as first demonstrated by Grosek and Kutz \cite{Kutz2014dmd}.

\subsection{Related work}
Bouwmans \cite{bouwmans2014traditional} or Sobral and Vacavant \cite{Sobralreview} provide recent and comprehensive reviews of methods for background modeling and related challenges. Among the many different techniques, the class of (robust) subspace models are prominent. PCA can be considered a traditional technique for describing the probability distribution of a static background. However, PCA has some essential shortcomings and many enhancements have been proposed since the method was first proposed for background subtraction by Oliver et al. \cite{oliver2000bayesian}, e.g. adaptive, incremental or independent PCA. A review of those traditional subspace models and related issues is provided by Bouwman \cite{bouwmans2009subspace}. While DMD is related to PCA and shares some of the same limitations, it can overcome others to greatly improve the performance. Grosek and Kutz \cite{Kutz2014dmd} have shown that DMD can be seen in fact as an approximation to robust PCA (see also \cite{bouwmans2015decomp}). The idea of RPCA is to separate a matrix $\mathbf{A}$ into a low-rank $\mathbf{L}$ and sparse component $\mathbf{S}$
\begin{equation}\label{eq:robustPCA}
\mathbf{A} = \mathbf{L} + \mathbf{S}
\end{equation}
This can be formulated as a convex optimization problem that minimizes a combination of the $l_{2}$ and $l_{1}$ norm. Applied to video data, the low-rank component describes the relatively static background environment, which is allowed to gradually change over time, while the second component captures the moving objects. This approach has gathered substantive attention for foreground detection since the idea was first introduced by Cand\`{e}s \cite{candes2011robust} - further extended by Zhou \cite{StableRPCA} for also capturing entry-wise noise. Bouwmans and Zahzah \cite{bouwmans2014robust} recently provided a comparative evaluation of the most prominent RPCA implementations, whose results show LSADM \cite{Goldfarb} and TFOCS \cite{Becker} algorithms perform best in extracting moving objects in terms of the F-measure. Guyon et al. \cite{guyon2012moving} show in detail how the former algorithm can be used for moving object detection. 

The problem formulation via RPCA leads to iterative algorithms with high computational costs. Most of the algorithms require repeated computation of the Singular Value Decomposition (SVD), so clearly the algorithms may be accelerated by using faster approximate SVD, aiming to find only the $k$ dominant singular values. Liu et al. \cite{Liu12limitedmemory} present a Krylov subspace-based algorithm for computing the first $k$ singular values with high precision. They showed that their LMSVD algorithm can reduce the computational time of RPCA substantially. Later they showed even greater computational savings with their Gauss–Newton method based SVD algorithm \cite{LiuGauss}. If high precision is not the main concern then approximate Monte-Carlo based SVD algorithms can be interesting alternatives \cite{frieze2004fast, drineas2006fast}. A different approach is via randomized matrix algorithms, which are surprisingly robust and provide significant speed-ups, while being simple to implement \cite{Martinsson201147}. Halko et al. \cite{halko2011rand} and Gu \cite{gu2015subspace} provide comprehensive surveys of randomized algorithms for constructing approximate matrix decompositions, while Mahoney \cite{Mahoney2011} gives a more general overview. One successful approximate robust PCA algorithm using a randomized matrix algorithms is given in GoDec \cite{Zhou11godec}.

\subsection{Motivation and contributions}
A core building block of the DMD algorithm, as for RPCA, is the SVD. As noted, traditional deterministic SVD algorithms are expensive to compute and with increasing data they often pose a computational bottleneck. We propose the use of a fast, probabilistic SVD algorithm, exploiting the rapidly decaying singular values of video data. Randomized SVD is a lean and easy to implement technique for computing a robust approximate low-rank SVD \cite{halko2011rand}. Compared to deterministic truncated or partial SVD algorithms, we gain computational savings in the order of 10 to 30 times. The next effect is to increase speed of about 2 to 3 times with randomized DMD, rather than deterministic SVD based DMD. Hence, randomized DMD may facilitate real-time processing of videos. Moreover, randomized SVD and DMD are embarrassingly parallel and we show that the computational performance can benefit from a Graphics Processing Unit (GPU) implementation. To demonstrate the applicability for motion detection, we have evaluated and compared dynamic mode decomposition on a comprehensive set of synthetic and real videos with other leading algorithms in the field.

The rest of this paper is organized as follows. Section \ref{sec:svd} presents randomized SVD as an approximation to the deterministic algorithms. Section \ref{sec:dmd} first introduces DMD and then shows how a low-rank DMD approximation using randomized SVD can be used for background modeling. Finally a detailed evaluation of DMD is presented in section \ref{sec:eval}. Concluding remarks and further research directions are given in section \ref{sec:conclusion}. 

\section{Singular Value Decomposition (SVD)}\label{sec:svd}
Matrix factorizations are fundamental tools for many practical applications in signal processing, statistical computing and machine learning. SVD is one such technique, used for data analysis, dimensionality reduction or data compression. Given an arbitrary real matrix $\mathbf{A} \in \mathbb{R}^{m\times n}$ we seek a decomposition, such that 
\begin{equation}\label{eq:svd}
\mathbf{A} = \mathbf{U}\mathbf{\Sigma}\mathbf{V}^{*}
\end{equation} 
where $\mathbf{U} \in \mathbb{R}^{m \times m}$ and $\mathbf{V} \in \mathbb{R}^{n \times n}$ are orthogonal matrices, and $\mathbf{\Sigma} \in \mathbb{R}^{m \times n}$ is a diagonal matrix with the same dimensions as $\mathbf{A}$ \cite{watkins2004fundamentals}. The columns of $\mathbf{U}$ and $\mathbf{V}$ are both orthonormal, called right and left singular vectors respectively. The singular values denoted as $\sigma_{i}$ are the diagonal elements of $\mathbf{\Sigma}$ sorted in decreasing order. While we assume a real matrix here, for generality we use the Hermitian transpose denoted as $^{*}$.

In practice we may be interested in a low-rank approximation of $\mathbf{A}$ with target rank $k \ll m,n$. Choosing the optimal target rank $k$ is highly dependent on the task, i.e. whether one is interested in a very good reconstruction of the original data or in a very low dimensional representation of the data. The reconstruction error for a low-rank approximation:
\begin{equation}\label{eq:lowrankapprox}
\| \mathbf{A} - \mathbf{U}_{k}\mathbf{\Sigma}_{k}\mathbf{V}^{*}_{k} \|_{F}=\sigma_{k+1}
\end{equation} 
is given by the singular value $\sigma_{k+1}$, where the index $F$ denotes the Frobenius norm. Thus, a reasonable small singular value gives a low reconstruction error, and we can denote $k$ in this case as the effective rank of the matrix $\mathbf{A}$. It can be proven that the exact low rank approximation is provided by the deterministic SVD, however the computational costs can be tremendous for large-scale problems, in particular for unstructured data. In the following, we present a faster randomized algorithm \cite{halko2011rand}.

\subsection{A Randomized SVD Algorithm}
Randomized matrix algorithms are approximate algorithms for linear algebra problems using random sampling and projections to accelerate the computation \cite{Mahoney2011}. Given an input matrix $\mathbf{A} \in \mathbb{R}^{m\times n}$ and a desired target rank $k \ll m,n$ the randomized algorithm for computing the approximate low-rank SVD can be roughly divided into two stages.

The first stage is concerned with finding a random low-dimensional subspace that best captures the column space of $\mathbf{A}$. Here the idea of random projections is used to build the basis for the column space. We simply draw $k$ random Gaussian vectors $\mathbf{x}_{i}$ and compute the following random sketch
 \begin{equation}\label{eq:samplevector}
 \mathbf{y}_{i} = \mathbf{A}\mathbf{x}_{i} \quad  \textrm{ for $i$=1,2,....,$k$ }
 \end{equation}
As a result from probability theory, it follows that the random vectors, and hence the set $\{\mathbf{y}_{i}\}$ are linearly independent. We can compute \eqref{eq:samplevector} more compactly as matrix-matrix product
\begin{equation}\label{eq:YAQ}
\mathbf{Y} = \mathbf{A}\mathbf{\Omega}
\end{equation}
where $\mathbf{\Omega} \in \mathbb{R}^{n \times k}$ is a random Gaussian matrix. We then compute the QR-Decomposition of $\mathbf{Y}$ to obtain the orthonromal matrix $\mathbf{Q} \in \mathbb{R}^{m\times k}$ so that
\begin{equation} \label{eq:QQA}	
\mathbf{A}  \approx  \mathbf{Q}\mathbf{Q}^{*}\mathbf{A}
\end{equation} 
is satisfied. 

In the second stage we project the input matrix $\mathbf{A}$ onto the low-dimensional subspace 
\begin{equation}
\mathbf{B} = \mathbf{Q}^{*}\mathbf{A}	
\end{equation} 
The action of the column space of $\mathbf{A}$ is now restricted to the relatively small (if $k \ll m,n$) matrix $\mathbf{B} \in \mathbb{R}^{k\times n}$. 
Subsequently we can cheaply compute the deterministic SVD of $\mathbf{B}$ as
\begin{equation} 
\mathbf{B} = \mathbf{\tilde{U}} \mathbf{\Sigma}\mathbf{V}^{*}	
\end{equation}  
The randomized algorithm can be justified as follows
\begin{equation} 
\begin{array}{cc ll ll} 	
\mathbf{A} 	& \approx & \mathbf{Q}\mathbf{Q}^{*}\mathbf{A}  \\
& \approx & \mathbf{Q}\mathbf{B} \\
& \approx & \mathbf{Q}\mathbf{\tilde{U}} \mathbf{\Sigma}\mathbf{V}^{*} \\
& \approx & \mathbf{U} \mathbf{\Sigma}\mathbf{V}^{*} \\
\end{array}	
\end{equation} 
Thus we can recover the right singular vectors by computing
\begin{equation} 
\mathbf{U} \approx \mathbf{Q}\mathbf{\tilde{U}}	
\end{equation}
Algorithm (\ref{RSVDalgorithm}) shows a prototype for computing the randomized SVD.\footnote{See supplementary materials for a more detailed algorithm and \textit{Python} implementation with oversampling parameter and subspace iterations.}
\begin{algorithm}[H]
	\caption{Randomized SVD (rSVD)}
	\label{RSVDalgorithm}
	\begin{algorithmic}[1]
		\Input $\mathbf{A} \in \mathbb{R}^{m \times n} $ and target rank $k$.
		\Require $m \geq n$, int $k \geq 1$ and $k \ll n$.
		\Procedure{rsvd}{$\mathbf{A}$, $k$}
		\State $\mathbf{\Omega} \gets$ rand(n, k)
		\Comment{Draw $n \times k$ random matrix.}
		\State $\mathbf{Y} \gets$ $\mathbf{A}* \mathbf{\Omega}$
		\Comment{Compute random sketch.}
		\State $\mathbf{Q} \gets$ qr($\mathbf{Y}$)
		\Comment{Economic QR-decomposition.}
		\State $\mathbf{B} \gets$ $\mathbf{Q}^{*} * \mathbf{A}$
		\Comment{Projection.}
		\State $\mathbf{\tilde{U}},s,\mathbf{V} \gets$ svd($\mathbf{B}$)
		\Comment{Deterministic SVD.}
		\State $\mathbf{U} \gets$ $\mathbf{Q} * \mathbf{\tilde{U}}$
		\Comment{Recover right singular vectors.}
		
		\Return $\mathbf{U} \in \mathbb{R}^{m \times k}, s \in \mathbb{R}^{k}, \mathbf{V} \in \mathbb{R}^{n \times k}$
		\EndProcedure
	\end{algorithmic}
\end{algorithm}
\begin{rmk}
	Common choices for generating the random matrix $\mathbf{\Omega}$  are the normal or uniform distribution.
\end{rmk}
The computational time can further reduced by first computing the QR-decomposition of $\mathbf{B}$ and then computing the SVD of the even smaller matrix $\mathbf{R} \in \mathbb{R}^{k\times k}$ (see Voronin et al. \cite{voronin2015rsvdpack} for further details).

The approximation error of a randomized SVD can be decreased by introducing a small oversampling parameter $p$. This means, instead of drawing $k$ random vectors, we generate $k+p$ samples, so that the likelihood of spanning the correct subspace is increased. A small oversampling parameter $p$ (e.g. $p=5$) is generally sufficient. Further, computing $q$ power iterations can increase the accuracy:
\begin{equation}
\mathbf{Y} = ( \mathbf{A}\mathbf{A}^{*} )^{q} \mathbf{A}\mathbf{\Omega}
\end{equation}
The power iterations drive the spectrum of $\mathbf{Y}$ down and the approximation error, which is proportional to the spectrum, decays exponentially with the number of iterations. Even if the signal-to-noise ratio is low, $q=1,2$ power iterations already achieve good results. For numerical reasons a practical implementation should use subspace iterations instead of power iterations \cite{gu2015subspace}. Halko et al. \cite{halko2011rand} showed that the approximation error of randomized SVD has the following error bound \cite{halko2011rand}, if the oversampling parameter is chosen equal to $k$, i.e. $l:=2k$
\begin{equation}
\mathbb{E}\big[\|\mathbf{A}-\mathbf{U}_{l} \mathbf{\Sigma}_{l}\mathbf{V}^{*}_{l}\|\big]=\sigma_{l+1} \Bigg[1+4\sqrt{\frac{2min(m,n)}{l-1}}\Bigg]^\frac{1}{2q+1}
\end{equation}
 
\subsection{Computational Costs}
SVD is often the bottleneck in practical large-scale applications. Many different methods for computing the SVD have been proposed and optimized for different problems, exploiting certain matrix properties. Thus, giving a detailed overview of the computational costs is difficult. 

In short however, the time complexity for the ordinary deterministic SVD algorithms is $O(mn^{2})$ if $m>n$, while modern partial SVD methods based on rank-revealing QR-factorization can reduce the time complexity to $O(mnk)$ \cite{demmel1997applied}. The randomized SVD algorithm using random sampling, as we have presented it here, needs two passes over the input matrix and also has asymptotic costs of $O(mnk)$. Hence, theoretically we have the same costs asymptotically - however from a practical point of view, it is much cheaper to compute a matrix-matrix multiplication than a column-pivoted QR factorization. The costs can be further deceased by exploiting certain matrix properties to compute a fast matrix-matrix multiplication of \eqref{eq:YAQ} to $O(mnlog(k))$ floating point operations. For example, the Subsampled Random Fourier Transform (SRFT) as proposed by Woolfe et al. \cite{woolfe2008fast} can be used.

In practice, the computational time of (randomized) SVD algorithms is also heavily driven by the computational platform used, the specific implementation and whether the matrix fits into the fast memory. An advantage of randomized SVD is that it can benefit from parallel computing. For example, permitting a GPU implementation, leading to dramatic acceleration \cite{voronin2015rsvdpack}. This is because the GPU architecture enables fast generation of random numbers and fast matrix-matrix multiplications.


\section{Dynamic Mode Decomposition (DMD)}\label{sec:dmd}
DMD is a data-driven method, fusing PCA with time-series analysis (Fourier transform in time) \cite{Kutz2013}. This integrated approach for decomposing a data matrix overcomes the PCA short-coming of performing an orthogonalization in space only. DMD is an emergent technique in the fluid mechanics community for analyzing the dynamics of non linear systems and was originally proposed by Schmidt \cite{schmid2010dynamic} and Rowley et al. \cite{rowley2009spectral}. Allowing the assessment of spatio-temporally coherent structures, with almost no underlying assumptions makes DMD interesting for video processing. Specifically, the resulting low-rank features are of interest for modeling the background of surveillance videos. In addition, DMD also allows predictions to be made about short-time future states of video streams \cite{kutz2015multi}. 

To compute the DMD, an ordered and evenly spaced data sequence describing a dynamical system is required. This applies naturally to videos, where a data matrix $\mathbf{D}\in \mathbb{R}^{m \times n}$ can be constructed so that the columns are $n$ consecutive grey coloured videos frames $\mathbf{f} \in \mathbb{R}^{m}$. The elements $d_{jt}$ of $\mathbf{D}$ refer to the intensity of a pixel in space ($j$) and time ($t$). As is common in the DMD literature, we denote such a data matrix also as snapshot sequence. Further, it is reasonable to assume that two consecutive video frames are related to each other in time. Mathematically, we can establish the following important relationship 
\begin{equation} 
\mathbf{f}_{t+1} = \mathbf{M} \mathbf{f}_{t}	
\end{equation} 
stating that there exists an unknown underlying linear operator $\mathbf{M} \in \mathbb{R}^{m \times m}$ that connects two consecutive video frames \cite{schmid2010dynamic}. Here the index $t \in \{1,2,...,n\}$ is denoting a frame in time.  It turns out that $\mathbf{M}$ is the Koopman operator whose eigenvalue decomposition describes the evolution of a video sequence \cite{tu2013dynamic}. Hence, the goal of DMD is to find an approximate decomposition of $\mathbf{M}$\footnote{Traditionally, the problem of obtaining the operator $\mathbf{M}$ was formulated in terms of a companion matrix in order to emphasize the deeper theoretical relationship to the Arnodli Algorithm and the Koopman operator. We refer to \cite{schmid2010dynamic, rowley2009spectral} for further theoretical details.}. It is also interesting to note that, while the operator $\mathbf{M}$ is considered to be linear, its eigenvectors and eigenvalues can also describe nonlinear dynamical systems. 

\subsection{Low-Rank Dynamic Mode Decomposition Algorithm}
To compute the DMD we proceed by first arranging the data matrix $\mathbf{D}\in \mathbb{R}^{m \times n}$ into two matrices: 
\begin{equation} 
\resizebox{.6\hsize}{!}{$
\mathbf{X} = \begin{bmatrix}
		& \mathbf{f}_{1} & | & \mathbf{f}_{2} & | & \mathbf{f}_{t} & | & ... & | & \mathbf{f}_{n-1}  
\end{bmatrix}  \in \mathbb{R}^{m \times (n-1)} 	
$}
\end{equation}
\begin{equation} 
\resizebox{.6\hsize}{!}{$
\mathbf{Y} = 	\begin{bmatrix}
& \mathbf{f}_{2} & | & \mathbf{f}_{3} & | & \mathbf{f}_{t+1} & | & ... & | & \mathbf{f}_{n} 
\end{bmatrix} \in \mathbb{R}^{m \times (n-1)}  
$}  	
\end{equation}
The left snapshot sequence $\mathbf{X}$ is approximately linked to the right sequence $\mathbf{Y}$ by the operator $\mathbf{M}$ as follows
\begin{equation} 
\mathbf{Y} \approx \mathbf{M} \mathbf{X}	
\end{equation}  
This is in fact a well known linear least squares problem
\begin{equation} 
min \| \mathbf{Y} - \mathbf{M}\mathbf{X} \|_{F}^{2}	
\end{equation} 
An estimate can be computed using the pseudo-inverse \cite{golub1965calculating} as follows
\begin{equation}\label{eq:pseudinverse} 
 \mathbf{M} = \mathbf{Y} \mathbf{X^{\dagger}} = \mathbf{Y} \mathbf{V}\mathbf{\Sigma^{-1}}\mathbf{U^{*}}	
\end{equation} 
where $\mathbf{U} \in \mathbb{R}^{m \times n} $ and $\mathbf{V} \in \mathbb{R}^{n \times n}$ are denoting the left and right singular values respectively, and $\mathbf{\Sigma} \in \mathbb{R}^{n \times m} $ the diagonal matrix with the corresponding singular values. However, this direct approach of computing the operator $ \mathbf{M}$ might not be feasible when dealing with high dimensional data, like videos. Instead it is more desirable to reduce the dimension first using a similarity transformation in order to find an approximate operator $\mathbf{\tilde{M}} \in \mathbb{R}^{n \times n}$ as
\begin{equation}\label{eq:simtrans}  
 \mathbf{\tilde{M}}=\mathbf{U}^{*}\mathbf{M}\mathbf{U}
\end{equation} 
In fact it can be shown that $\mathbf{M}$ and $\mathbf{\tilde{M}}$ have the same eigenvalues \cite{demmel1997applied}. Using the similarity transformation draws a connection between DMD and PCA by projecting $\mathbf{M}$ onto the principal components (left singular vectors) $\mathbf{U}$. We obtain $\mathbf{\tilde{M}}$ by plugging \eqref{eq:pseudinverse} into \eqref{eq:simtrans} as follows (note that $\mathbf{U}$ is a matrix with orthonormal columns and hence $\mathbf{U}^{*}\mathbf{U}=\mathbf{I}$)
\begin{equation} 
\mathbf{\tilde{M}} = \mathbf{U}^{*}\mathbf{M}\mathbf{U} = \mathbf{U^{*}\mathbf{Y}\mathbf{V}\mathbf{\Sigma^{-1}}}	 
\end{equation}  
It can be seen that the SVD plays a central role in computing the DMD. Computing the SVD can be computationally expensive, however exploiting the low-dimensional structure of video data (with rank $k \ll n$) allows us to use fast approximate low-rank decomposition techniques, e.g., randomized SVD (rSVD) as described in Section \ref{sec:svd}. We denote this approach using rSVD for computing an approximate low-rank dynamic mode decomposition with a specified target rank $k$, as randomized DMD (rDMD). In this case, the dimension of the linear operator reduces to $\mathbf{\tilde{M}} \in \mathbb{R}^{k \times k} $. The structure of $\mathbf{\tilde{M}}$ is revealed by computing the eigenvalue decomposition of $\mathbf{\tilde{M}}$ as
\begin{equation} 
\mathbf{\tilde{M}}\mathbf{W} =\mathbf{\Lambda} \mathbf{W}	
\end{equation} 
where $\mathbf{W} \in \mathbb{C}^{k \times k}$ is the eigenvector matrix and $\mathbf{\Lambda} \in \mathbb{C}^{k \times k}$ is a diagonal matrix containing the eigenvalues $\lambda$. 
The dynamic modes $\mathbf{\Phi} \in \mathbb{C}^{m \times k}$ are then computed by relating the eigenvectors back to $\mathbf{M}$ as either \cite{schmid2010dynamic}
\begin{equation} 
\mathbf{\Phi}=[\mathbf{\phi_{1}}, ... , \mathbf{\phi_{k}}]=\mathbf{U}\mathbf{W}	
\end{equation} 
or more generally as \cite{tu2013dynamic}
\begin{equation} 
\mathbf{\Phi}=\mathbf{Y}\mathbf{V}\mathbf{\Sigma^{-1}}\mathbf{W}	
\end{equation} 
We favor the latter approach.

\begin{figure}[htp]
	\centering
	\DeclareGraphicsExtensions{.eps}
	\includegraphics[width=0.6\textwidth]{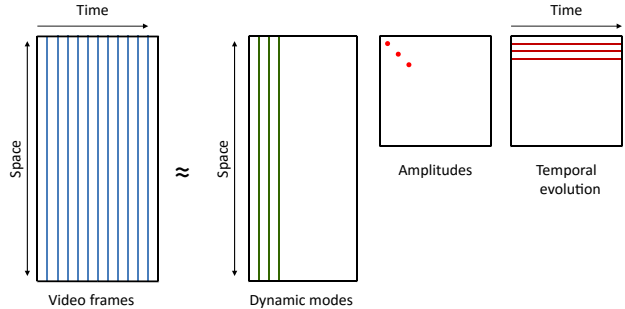}
	\caption{Illustration of low-rank dynamic mode decomposition.}
	\label{fig:DMDdecomp}
\end{figure}

The original data matrix $\mathbf{D}$ can be reconstructed by noting that the snapshots can be represented as the linear combination \cite{jovanovic2012low}
\begin{equation}\label{eq:reconstructDMD} 
\mathbf{f}_{t} \approx \sum_{i=1}^{k} b_{i}\mathbf{\phi}_{i}\lambda_{i}^{t-1}	
\end{equation}
where $\lambda_{i}$ denotes the $i$th eigenvalue, ${\phi_{i}}$ the $i$th dynamic mode and $b_{i}$ the corresponding amplitude. Since $b_{i}$ is time independent $\mathbf{f}_{1}$ reduces to 
\begin{equation}\label{eq:amplitudes} 
\mathbf{f}_{1} \approx \sum_{i=1}^{k} b_{i}\mathbf{\phi}_{i}=\mathbf{\Phi}\mathbf{b}	
\end{equation}
The parameter vector $\mathbf{b} \in \mathbb{C}^{k}$ can be estimated by the linear least squares method \cite{Kutz2014dmd}.
Figure~\ref{fig:DMDdecomp} illustrates how the approximate low-rank DMD can be expressed as 
\begin{equation}\label{eq:dmddecomposition} 
\mathbf{D} \approx \mathbf{\Phi}\mathbf{B}\mathbf{V_{and}}	
\end{equation} 
where $\mathbf{B}\in \mathbb{C}^{k \times k}$ is a diagonal matrix of the amplitudes
\begin{equation} 
		\mathbf{B}= \begin{pmatrix}
			b_{1} 	&       &         	&  		\\
			 	  	& b_{i} &  			&  		\\
			 	  	&       & \ddots 	&   	\\
			 		&  		&  			& b_{k}
		\end{pmatrix}
\end{equation} 
and $\mathbf{V_{and}} \in \mathbb{C}^{k \times n} $ is the Vandermonde matrix of the eigenvalues 
\begin{equation} 
		\mathbf{V_{and}}= \begin{pmatrix}
			1 		& \lambda_{1} & \cdots & \lambda_{1}^{n-1}  \\
			1		& \lambda_{2} & \cdots & \lambda_{2}^{n-1} 	\\
			\vdots	& \vdots  & \ddots & \vdots  		\\
			1		& \lambda_{k} & \cdots & \lambda_{k}^{n-1}
		\end{pmatrix}
\end{equation}
From the Vandermonde matrix it is clear that temporal dynamics, retrieved by the DMD, consists of single (distinct) frequencies.

The prototype Algorithm (\ref{randDMDalgorithm}) summarizes the method for computing the DMD using rSVD.
\begin{algorithm}[htp]
\caption{Randomized DMD (rDMD)}
\label{randDMDalgorithm}
	\begin{algorithmic}[1]
		\Input $\mathbf{D} \in \mathbb{R}^{m \times n}$ and target rank $k$.
		\Require $m \geq n$, integer $k \geq 1$ and $k \ll n$
		\Procedure{rDMD}{$\mathbf{D}$, $k$}
			\State $\mathbf{X,Y} \gets$ $\mathbf{D}$
			\Comment{Left/right snapshot sequence.}
			\State $\mathbf{U},\mathbf{s},\mathbf{V} \gets$ rsvd($\mathbf{X}$, k)
			\Comment{Low-rank rSVD.}
			\State $\mathbf{S} \gets$ $diag(\mathbf{s}^{-1})$
			\Comment{Diagonal matrix.}
			\State $\mathbf{M} \gets$ $\mathbf{U}^{*} * \mathbf{Y} * \mathbf{V} * \mathbf{S}$
			\Comment{Least squares fit.}
			\State $\mathbf{W},l \gets$ $eig(\mathbf{M})$
			\Comment{Eigenvalue decomposition.}
			\State $\mathbf{F} \gets$ $\mathbf{Y} * \mathbf{V} * \mathbf{S} * \mathbf{W}$
			\Comment{Compute modes $\mathbf{\Phi}$.}
			\State $\mathbf{b} \gets$ $lstsq(\mathbf{F},\mathbf{x}_{1})$
			\Comment{Compute amplitudes.}
			\State $\mathbf{V} \gets$ $vander(l)$
			\Comment{Vandermonde matrix.}
		\State \textbf{return} $\mathbf{F} \in \mathbb{C}^{m \times k}, \mathbf{b} \in \mathbb{C}^{k}, \mathbf{V} \in \mathbb{C}^{k \times n}$
		\EndProcedure
	 \end{algorithmic}
\end{algorithm}

\subsection{DMD for Background Modeling}
\begin{figure}[htp]
	\centering
	\DeclareGraphicsExtensions{.eps}
	\includegraphics[width=0.7\textwidth]{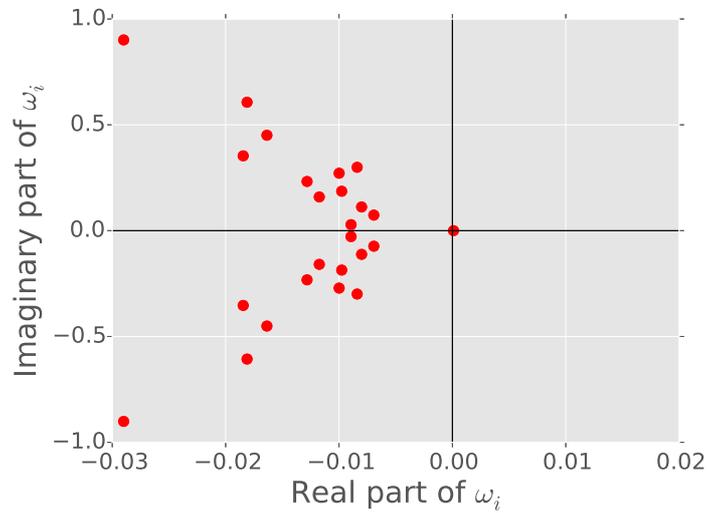}
	\caption{Fourier modes corresponding to a low-rank dynamic mode decomposition.}
	\label{fig:omega}
\end{figure}

\begin{figure}[htp]
	\centering
	\begin{subfigure}[t]{0.3\textwidth}
		\centering
		\DeclareGraphicsExtensions{.eps}
		\includegraphics[height=1.3in]{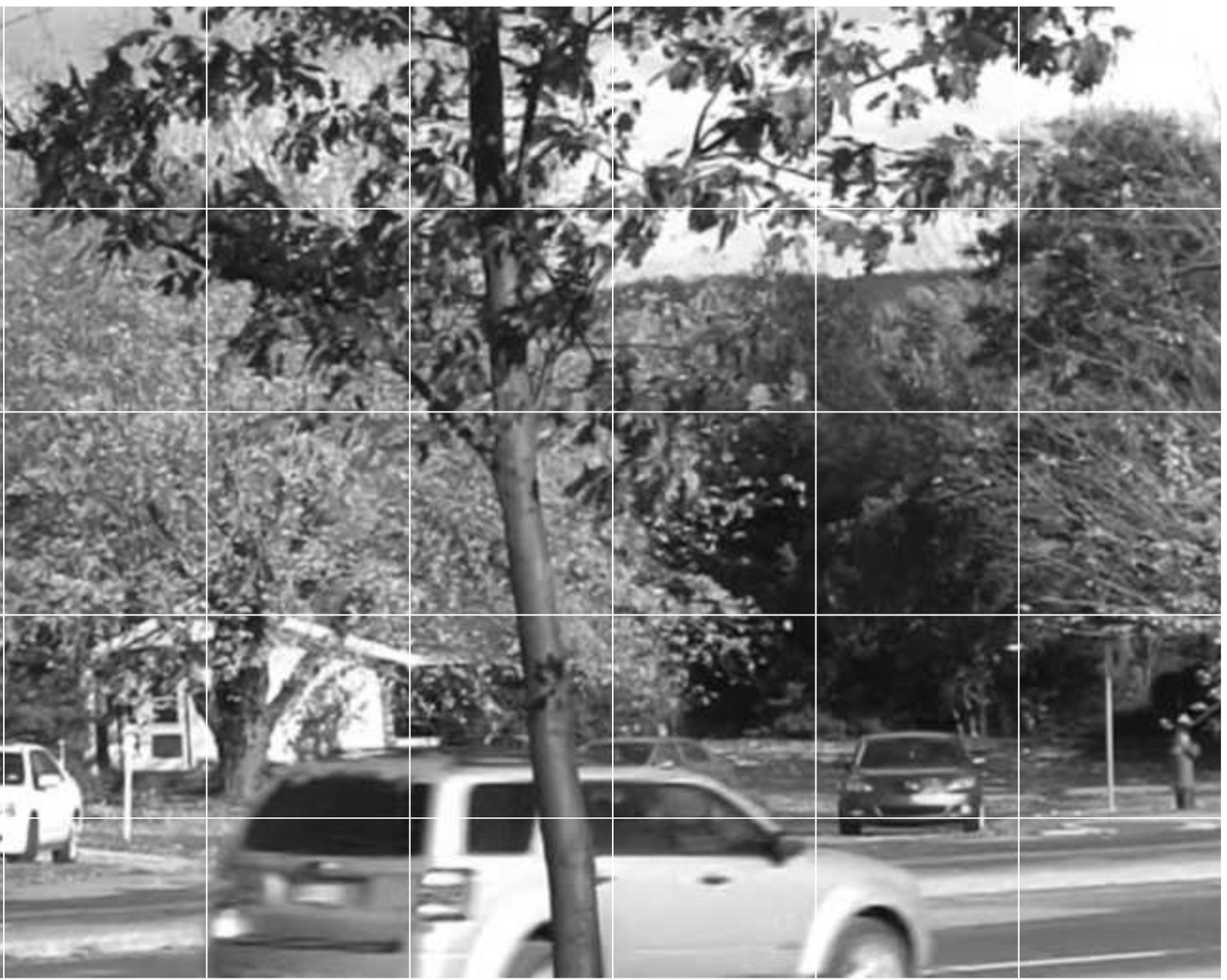}
		\caption{Original frame.}
	\end{subfigure}
	~~
	\begin{subfigure}[t]{0.3\textwidth}
		\centering
		\DeclareGraphicsExtensions{.eps}
		\includegraphics[height=1.3in]{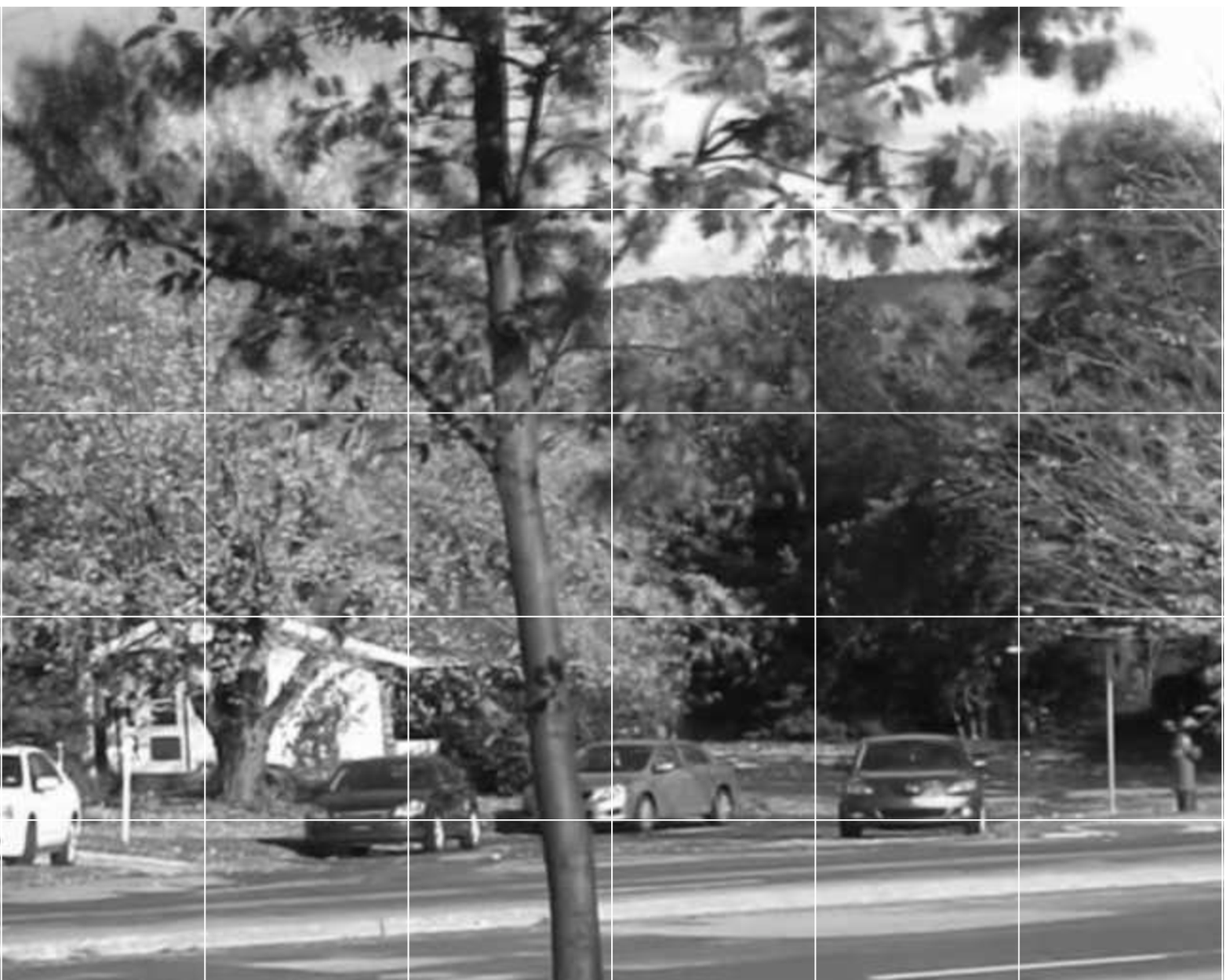}
		\caption{Zero mode.}
	\end{subfigure}
	
	\begin{subfigure}[t]{0.3\textwidth}
		\centering
		\DeclareGraphicsExtensions{.eps}
		\includegraphics[height=1.3in]{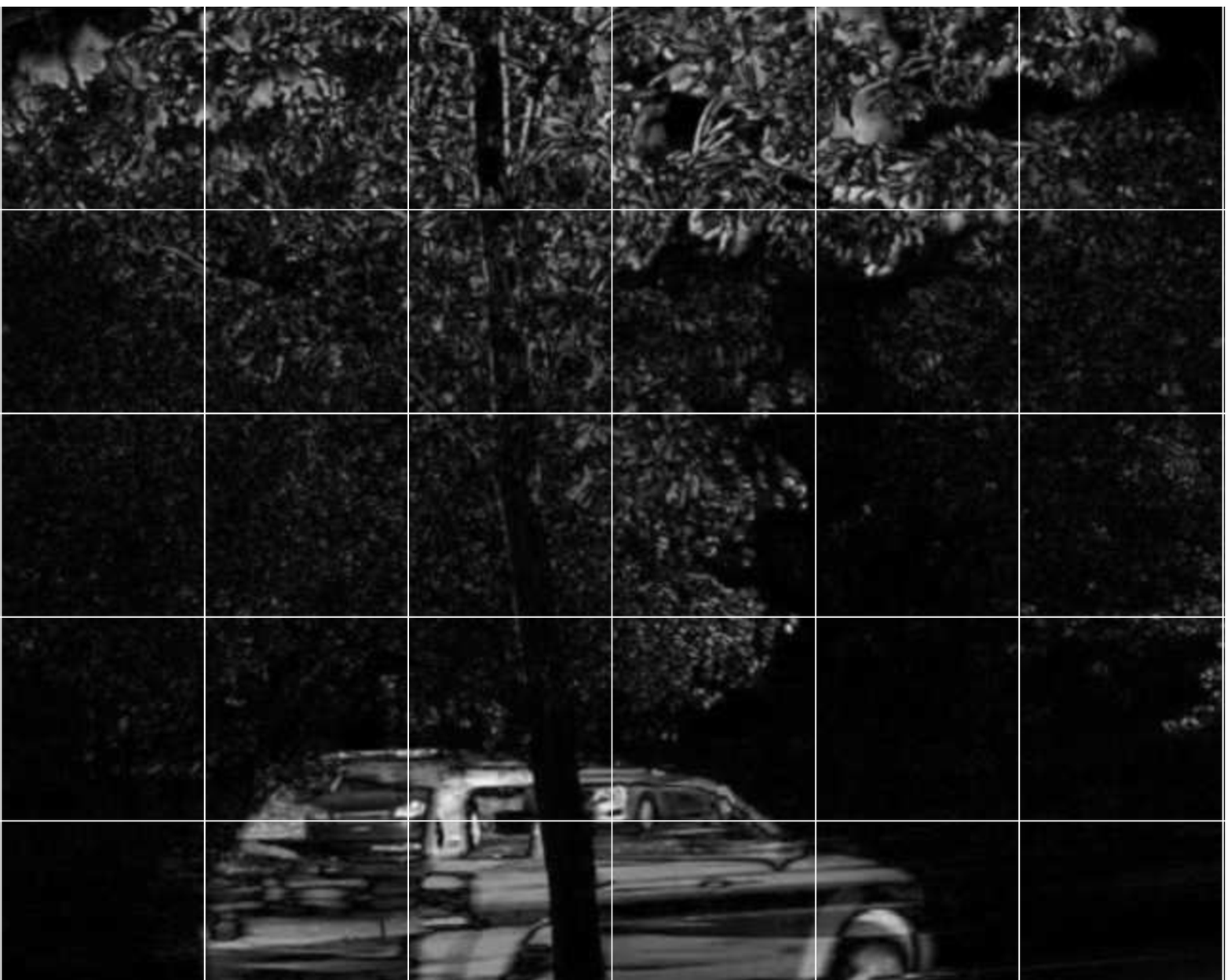}
		\caption{Foreground reconstructed with zero mode only.}
	\end{subfigure}
	~~
	\begin{subfigure}[t]{0.3\textwidth}
		\centering
		\DeclareGraphicsExtensions{.eps}
		\includegraphics[height=1.3in]{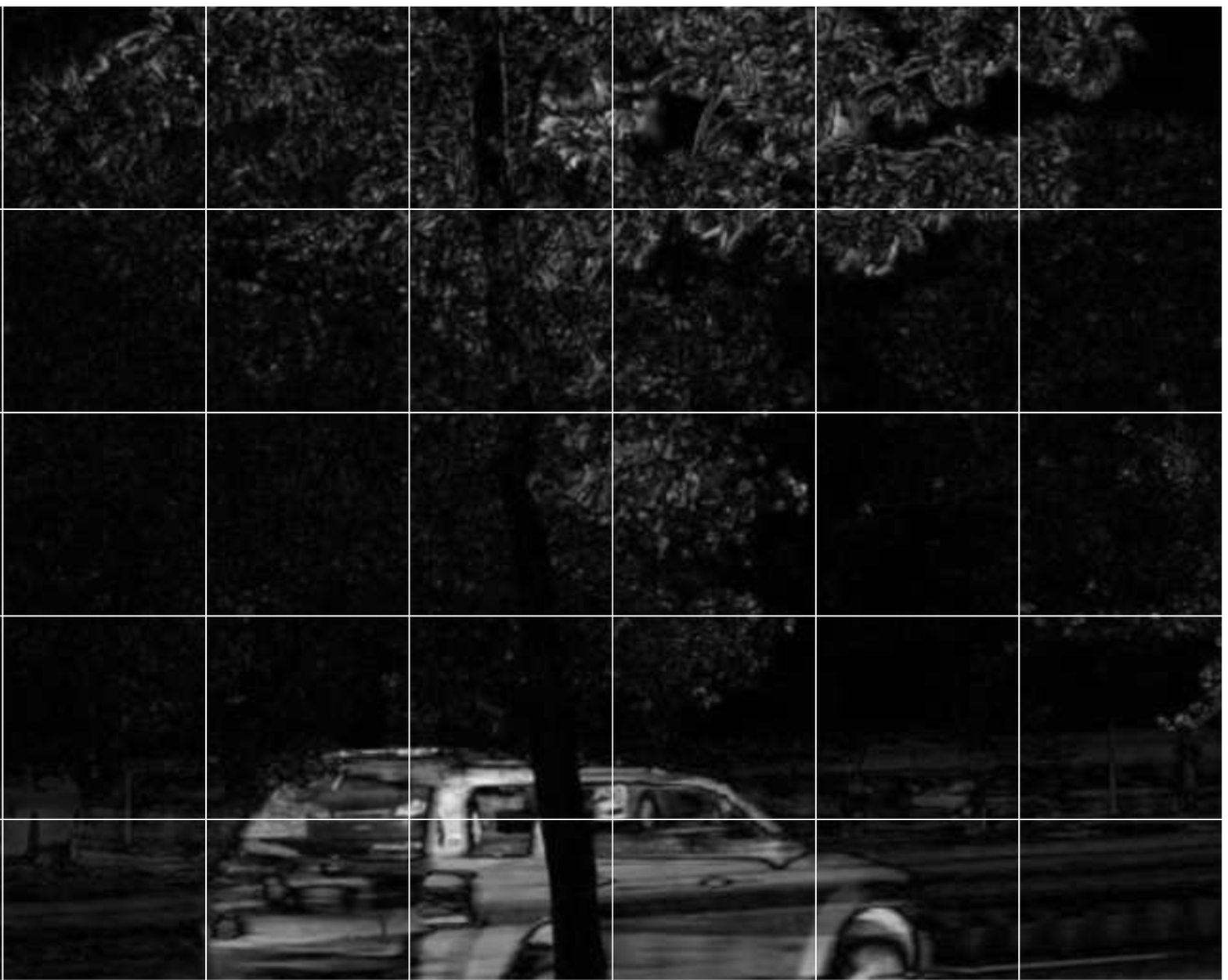}
		\caption{Foreground reconstructed with the 3 smallest modes.}
	\end{subfigure}
	\caption{Illustration of background modeling using DMD.}
	\label{fig:modes}
\end{figure}

In the previous section we have seen how DMD can be used to decompose and reconstruct a matrix. However, using \eqref{eq:dmddecomposition} for modeling the video background directly is a bad strategy. Of course, we can hope that when computing the low-rank dynamic mode decomposition, that the dominant dynamic modes are not corrupted by any moving objects and only capture background structures. Like PCA, this works when we train DMD on a set of clean video frames. However, that is an unrealistic scenario in real world applications. More desirable is a decomposition into low-rank $\mathbf{L}$ (background components) and sparse components $\mathbf{S}$ (foreground components) similar to RPCA \cite{candes2011robust}
\begin{equation}
\mathbf{D} = \mathbf{L} + \mathbf{S}
\end{equation}
Unlike robust PCA, DMD is not capable of directly separating a matrix into these two components. Instead, DMD allows us to compute an approximation to it. First let us connect the DMD eigenvalues $\lambda$ to the Fourier modes $\omega$ as follows \cite{Kutz2014dmd}
\begin{equation}
 \omega_{i} = \frac{ln(\lambda_{i})}{\Delta t}
\end{equation}
For standard videos we simply assume the time step $\Delta t=1$ and hence $\omega_{i}=ln(\lambda_{i})$. By construction, the eigenvalues are complex. Hence the Fourier modes allow us to reveal interesting properties about the relating dynamic modes. The real part of $\omega$ determines the mode's evolution over time, while the imaginary part is related to the mode's oscillations. Now let us rewrite \eqref{eq:dmddecomposition} in terms of the Fourier modes for a $k$ low-rank decomposition of a video matrix
\begin{equation}\label{eq:dmdfourier} 
\mathbf{D} \approx \mathbf{\Phi}\mathbf{B}\mathbf{V_{and}} = \sum_{i=1}^{k} b_{i}\mathbf{\phi}_{i} \exp(\omega_{i} \mathbf{t})	
\end{equation} 
where $\mathbf{t}=[0,1,...,(n-1)]$ is the time vector. From \eqref{eq:dmdfourier} it is clear that the Fourier modes dictate how the modes evolve, i.e., decay or grow in time. In light of this, the set of $k$ modes $\{\mathbf{\phi}_{i}\}$ can be separated into a set that contains only Fourier modes $\{w_{l} : \|w_{l}\| \ll 1\}$ who evolve slowly over time and corresponds to background modes. The second set $\{w_{s}\}$ contains modes describing fast moving objects. Exploiting this, \eqref{eq:dmdfourier} can be rewritten as
\begin{equation} 
\mathbf{D} \approx \sum_{i\in l} b_{i}\mathbf{\phi}_{i} \exp(\omega_{i} \mathbf{t})	+ \sum_{i \in s} b_{i}\mathbf{\phi}_{i} \exp(\omega_{i} \mathbf{t})
\end{equation} 
The background video can then be reconstructed as follows
\begin{equation}
\mathbf{L} = \sum_{i \in l} b_{i}\mathbf{\phi}_{i} \exp(\omega_{i} \mathbf{t})	
\end{equation} 
Foreground objects (sparse components) can be identified as difference between the original video data and the background video $\mathbf{L}$ (discarding the imaginary part)
\begin{equation}
\mathbf{S} = \|\mathbf{D} - \mathbf{L} \|_{2}	
\end{equation} 
We illustrate the concept on a real video in the following examples. Figure~\ref{fig:omega} shows the Fourier modes of a low-rank dynamic mode decomposition with target rank $k=25$. 
The Fourier mode $\|\omega_{0}\|\approx 0$ identifies the background mode, shown in Figure~\ref{fig:modes} (b). However, using just the zero mode leads to a static background model. Figure~\ref{fig:modes} (c) shows that the waving tree is captured as foreground object when using the zero mode only. Hence, to better cope with dynamic backgrounds, it is favorable to select a subset of modes $\|w_{b}\| \ll 1$ for background modeling. Using the first 3 modes decreases the false positive rate, as shown in figure~\ref{fig:modes} (d). Deciding upon the number of modes used for modeling the background was semi-arbitrary and we achieved qualitatively good results with 3 to 5 modes --- whereas using the zero mode is computationally faster.   

\section{Experimental Evaluation}\label{sec:eval}

In this section we evaluate both the accuracy and computational performance of the proposed algorithm and compare it to other state-of-the-art methods. To evaluate the effectiveness of rDMD for detecting moving objects we use two benchmark datasets. First, we test rDMD on ten synthetic videos from the BMC 2012 (Background Models Challenge) dataset \cite{vacavant2013benchmark}. Further, to evaluate the performance on real videos, we use eight videos from the ChangeDetection.net (CD) dataset \cite{wang2014cdnet}. The selected videos represent challenging examples in motion detection. For example:

\begin{itemize}
	\item \textbf{Bootstrapping}: A sequence of clean background images which are not available for training. 
	\item \textbf{Dynamic backgrounds}: Moving objects which belong to the background like waving trees, rain or snowfall. 
	\item \textbf{Illumination changes}: Gradual illumination changes of the environment due to fog or sun.
	\item \textbf{Camouflage}: Foreground objects which have the same pixel intensity as background elements, i.e. same color. 
\end{itemize}

\subsection{Evaluation Measures}
To evaluate the performance of background subtraction algorithms, a binary foreground mask using a suitable distance measure $d(\cdot)$ has to be computed
\begin{equation}\label{eq:thres}
\mathcal{X}_{t}(j) = \left \{
\begin{array}{lc lc} 	
1 & \quad  \textrm{if $d( f_{jt}-b_{jt}) > \tau$} \\
0 & \quad  \textrm{otherwise}	
\end{array}
\right.
\end{equation}
For instance, the Euclidean distance is a common choice for measuring the distance between pixels of the actual video and the re-constructed background frame \cite{benezeth2010comparative}. However, more sophisticated measures can be formulated and allow adaptive thresholding. The resulting vector $\mathcal{X}_{t}$ is called the \emph{foreground} or \emph{motion} mask and its elements are binary $x\in\{0,1\}$. In the outcomes, $1$ classifies a pixel belonging to a foreground object, otherwise $0$ as a background element. Thus we can visualize the classification results as an confusion matrix 
\begin{equation}
\scalebox{0.8}{
\begin{tabular}{c c c c |c}
	& 	& \multicolumn{2}{ c }{\textbf{Truth }}  \\ 
		& \multicolumn{1}{ c| }{} & 0 & 1  &    \\ 
	\cline{2-5}
	\multicolumn{1}{c}{\multirow{2}{*}{\textbf{Prediction}} } &
	\multicolumn{1}{ c| }{0} & TN & FN & \#pred neg    \\
	& \multicolumn{1}{ c| }{1} & FP & TP & \#pred pos   \\ \cline{2-5}
	& \multicolumn{1}{ c| }{} & \#true neg & \#true pos &   \\ 
\end{tabular}
}
\end{equation}
where TP denotes the (number of) True Positive predictions, i.e. pixels which are correctly classified as belonging to a moving foreground object. Similarly TN denotes the (number of) True Negative predictions, i.e. pixels which are correctly classified as background. False Positive (FP) and False Negative (FN) are the respective misclassifications for foreground and background elements. Based on the confusion matrix we can compute the following evaluation measures.

Recall (also called sensitivity, true positive rate or hit rate) measures the algorithm's ability to correctly detect pixels belonging to foreground objects. It is computed as the ratio of predicted true positives to the total number of true positive foreground pixels
\begin{equation}
\mathbf{Recall} = \frac{\mathrm{TP}}{\mathrm{TP+FN}}
\end{equation} 

Precision (also called false alarm rate or true positive accuracy) measures how confident we can be that a positive classified pixel actually belongs to a foreground object. It is computed as the ratio of predicted true positives to the total number of pixels predicted as foreground objects 
\begin{equation}
\mathbf{Precision} = \frac{\mathrm{TP}}{\mathrm{TP+FP}}
\end{equation}

Specificity (also called true negative rate) measures the algorithm's ability to correctly predict pixels belonging to the background. It is computed as the ratio of true negatives to the total number of true negative foreground pixels
\begin{equation}
\mathbf{Specificity} = \frac{\mathrm{TN}}{\mathrm{TN+FP}}
\end{equation}

The F-measure combines recall and precision as their harmonic mean, weighting both measures evenly, defined as
\begin{equation}
\mathbf{F} = 2\times\frac{\mathrm{Recall \times Precision}}{\mathrm{Recall+Precision}}
\end{equation} 
More general definitions of the F-measure also allow different weighting schemes.  

From~\eqref{eq:thres} it is obvious that the classification results depend on a pre-defined fixed threshold $\tau$. To get a global understanding of the algorithm's behaviors the evaluation measures can be computed over a range of different thresholds. The results can then be visualized using precision-recall and Receiver Operator Characteristics (ROC) curves. An advantage of ROC graphs, plotting precision vs 1-specificity, is the insensitivity to changes in the class distribution \cite{fawcett2005ROC}. In particular in dynamic environments such as videos, the number of pixels belonging to foreground objects can vary significantly over frames and is generally much less then the number of pixels belonging to the background. A further advantage of using ROC curves is the convenient way to summarize the performance with a global single scalar value measuring the Area Under the Curve (AUC). The perfect ROC curve has an AUC of 1, while random guessing yields an AUC of $0.5$. Thus a method with an AUC close to 0.5 or below can be considered as useless, while a method with a higher AUC is preferred. 

\subsection{Results}
DMD is formulated as a batch algorithm here, i.e, previous modeled sequences do not effect the following. This allows the algorithm to adapt to changes in the scene, e.g., illumination changes. Also, foreground objects that become background objects (like a recently parked car) can be better captured. On the other hand, it does not allow for dealing with `sleeping' foreground objects. The performance varies with the number of modes and the length of the snapshot sequence. Our results show that a snapshot length of about 100 to 300 video frames can be separated with a very low number of modes, e.g. $k \in \{9,11,...,15\}$. If the video is less noisy, a lower number of dynamic modes is sufficient. However, depending on how fast the foreground objects are moving, using less than 100 frames often leads to a poor detection performance. Another important issue is the choice of the initial condition used for computing the amplitudes. The default option is to use the first frame of the sequence, as stated in \eqref{eq:amplitudes}, however we often achieved better results using the median frame instead. Another interesting option is to recompute the amplitudes for small chunks of the sequence. This allows better capture of sudden illumination changes.  

For the rDMD algorithm, two further tuning parameters $p$ and $q$ can be specified. The former is the oversampling parameter and the latter controls the number of power iterations of the rSVD algorithm. For computing rDMD in the following we keep the two parameters fixed as $p=2$ and $q=1$. This parameter setting recovers almost exactly the results achieved with the ordinary DMD algorithm.

\begin{figure*}[htp]
	\centering
	\begin{subfigure}{0.85\textwidth}
		\DeclareGraphicsExtensions{.eps}
		\includegraphics[width=1\textwidth]{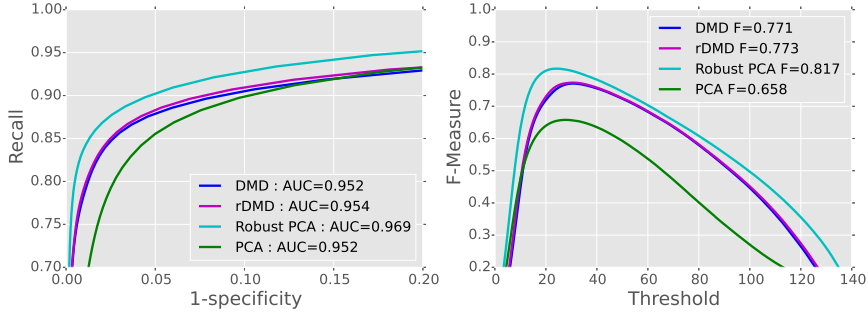}
		\caption{CD baseline video 'Highway' frame 500 to 700.}
	\end{subfigure}

	\begin{subfigure}{0.85\textwidth}
		\DeclareGraphicsExtensions{.eps}
		\includegraphics[width=1\textwidth]{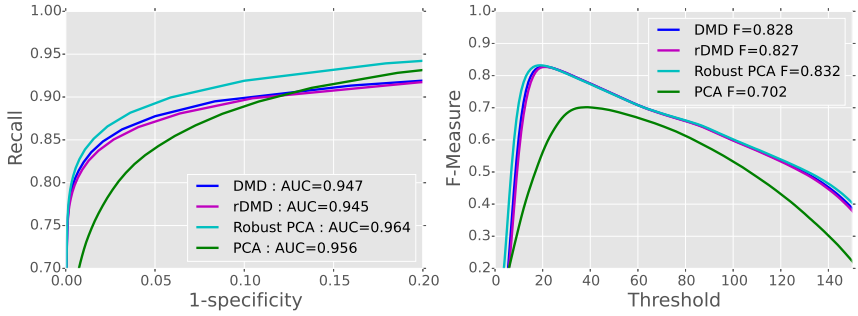}
		\caption{CD baseline video 'Pedestrians' frame 600 to 800.}
	\end{subfigure}
	\caption{Performance evaluation of rDMD, DMD, PCA and RPCA on two videos. The left column presents ROC curves and the right column the F-measure. While RPCA performs best, DMD/rDMD can be seen as a good approximation.}
	\label{fig:roc}
\end{figure*}

We first illustrate in Figure~\ref{fig:roc} the performance of rDMD compared to ordinary DMD, PCA \cite{oliver2000bayesian} and robust PCA \cite{candes2011robust} on two videos. While RPCA performs best both in terms of the AUC and the F-measure, rDMD and DMD can be seen as a reasonable approximation. The results also show that the performance difference between rDMD and DMD is insignificant. As expected, DMD performs significantly better than PCA in terms of the F-measure.

Table~\ref{Tab:evaluationBMC} shows the results of randomized DMD for the ten synthetic videos of the BMC dataset and compares them with three leading robust PCA algorithms: LSADM, TFOCS and GoDec. LSADM \cite{Goldfarb} is a principal component pursuit algorithm, while TFOCS \cite{Becker} is a quantization-based principal component pursuit algorithm. GoDec \cite{Zhou11godec} is an approximated RPCA algorithm based on bilateral random projections and like rDMD uses the concept of randomized matrix algorithms. Overall the average F-measure shows that the detection performance of rDMD is about 4$\%$ lower than the RPCA algorithms. The slightly poorer performance is due to the Street 512 and Rotary 522 videos, emulating windy scenes with additional noise. In these cases the background is very dynamic and the precision of the DMD algorithm is decreased. However, this problem can be compensated for by post-processing the obtained foreground mask with a median filter. The overall performance of this approach leads to an improvement of about 2$\%$. The results on the other videos show that DMD is flexible enough to deal with illumination changes like clouds, fog or sun. 
\begin{table*}[ht]
	\centering
	\scalebox{0.7}{
		\begin{tabular}{ l l c c c c c c c c c c c} 
			\hline
			& \multicolumn{1}{l}{Measure}
			& \multicolumn{5}{c}{Street} & \multicolumn{5}{c}{Rotary} 
			& \multicolumn{1}{c}{Average}
			\\
			\cmidrule(r){3-7}
			\cmidrule(r){8-12}
			&  				& 112   & 212   & 312 & 412  & 512  & 122   & 222 & 322 & 422  & 522\\
			\hline
			
			\multirow{3}{*}{\rotatebox[origin=c]{0}{ \parbox{3cm}{LSADM \\ Goldfarb et al. \cite{Goldfarb} }     }} 
			& Recall 		& 0.874 & 0.857 & 0.906 & 0.862 & 0.840 & 0.878 & 0.880 & 0.892 & 0.782 & 0.830 & -
			\\
			& Precision    	& 0.830 & 0.965 & 0.867 & 0.935 & 0.742 & 0.940 & 0.938 & 0.892 & 0.956 & 0.869 & -
			\\
			& F-Measure 	& 0.851 & 0.908 & 0.886 & 0.897 & 0.788 & 0.908 & 0.908 & 0.892 & 0.860 & 0.849 & 0.880 
			\\ \hline
			
			\multirow{3}{*}{\rotatebox[origin=c]{0}{ \parbox{3cm}{TFOCS \\ Becker et al. \cite{Becker}}   }} 
			& Recall 		& 0.910 & 0.843 & 0.867 & 0.903 & 0.834 & 0.898 & 0.892 & 0.892 & 0.831 & 0.877 & -
			\\
			& Precision    	& 0.830 & 0.965 & 0.899 & 0.889 & 0.824 & 0.924 & 0.932 & 0.887 & 0.940 & 0.879 & -
			\\
			& F-Measure 	& 0.868 & 0.900 & 0.882 & 0.896 & 0.829 & 0.911 & 0.912 & 0.889 & 0.882 & 0.878 & 0.885 
			\\ \hline
			 
			\multirow{3}{*}{\rotatebox[origin=c]{0}{ \parbox{3cm}{GoDec \\ Zhou and Tao \cite{Zhou11godec}}   }} 
			& Recall 		& 0.841 & 0.875 & 0.850 & 0.868 & 0.866 & 0.822 & 0.879 & 0.792 & 0.813 & 0.866 & -
			\\
			& Precision    	& 0.965 & 0.942 & 0.968 & 0.948 & 0.902 & 0.900 & 0.921 & 0.953 & 0.750 & 0.837 & -
			\\
			& F-Measure 	& 0.899 & 0.907 & 0.905 & 0.906 & 0.884 & 0.859 & 0.900 & 0.865 & 0.781 & 0.851 & 0.876 
			\\ \hline						
			
			\multirow{3}{*}{\rotatebox[origin=c]{0}{ \parbox{3cm}{ rDMD \\ } }} 
			& Recall 		& 0.873 & 0.855 & 0.760 & 0.805 & 0.783 & 0.883 & 0.860 & 0.772 & 0.800 & 0.834 & -
			\\
			& Precision    	& 0.887 & 0.912 & 0.902 & 0.900 & 0.656 & 0.896 & 0.907 & 0.876 & 0.902 & 0.770 & -
			\\
			& F-Measure 	& 0.880 & 0.882 & 0.825 & 0.850 & 0.714 & 0.889 & 0.882 & 0.820 & 0.848 & 0.800 & 0.839
			\\ \hline
			
			\multirow{3}{*}{\rotatebox[origin=c]{0}{ \parbox{3cm}{rDMD \\ (with median filter) }   }} 
			& Recall 		& 0.859 & 0.833 & 0.748 & 0.793 & 0.801 & 0.862 & 0.834 & 0.808 & 0.761 & 0.831 & -
			\\
			& Precision    	& 0.906 & 0.935 & 0.924 & 0.916 & 0.879 & 0.922 & 0.936 & 0.892 & 0.941 & 0.894 & -
			\\
			& F-Measure 	& 0.882 & 0.881 & 0.826 & 0.850 & 0.838 & 0.891 & 0.882 & 0.847 & 0.842 & 0.861 & 0.860
			\\ \hline			
			
			\hline
		\end{tabular}
	}
	\caption{Evaluation results of ten synthetic videos from the BMC dataset. For comparison, the results of three other leading RPCA algorithms are presented, adapted from \cite{bouwmans2014robust}.}
	\label{Tab:evaluationBMC}
\end{table*}

\begin{figure}[htp]
	\centering
	\DeclareGraphicsExtensions{.eps}
	\includegraphics[width=0.55\textwidth]{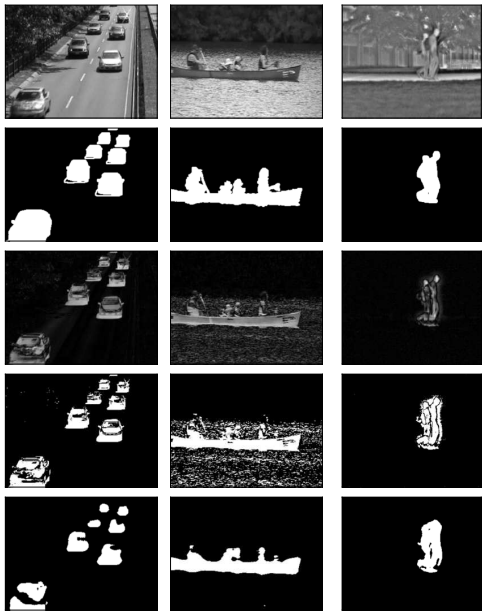}
	\caption{Visual results for 3 example frames from the CD Videos: Highway, Canoe and Park. The top row shows the original gray scaled image and the second row the corresponding true foreground mask. The third line shows the differencing between the reconstructed background and the original frame. Rows four and five are the thresholded and median filtered foreground masks, respectively.   }
	\label{fig:results}
\end{figure}

We show in Table~\ref{Tab:evaluationCD} the evaluation results of 8 real videos from the CD dataset. The videos are from three different categories: `Baseline', `Dynamic Background' and `Thermal'. At first glance the overall performance of the rDMD algorithm looks poor. This can be related to several challenges faced here. While the performance on the two baseline videos 'Highway' and 'Pedestrians' are good, issues arise for the other two baseline videos. The PETS2006 is difficult, due to camouflage effects for the DMD algorithm, as well as because some of the objects are sleeping foreground objects. The `Office' shows even more drastically that DMD cannot cope with sleeping foreground objects. However, integrating DMD into a simple system allowing for background maintenance can help to overcome this problem. The two dynamic background videos show the same problem as before with the synthetic videos. The recall rate is excellent, while the precision is lacking. But again just using a simple median filter for pre-processing increases the performance greatly. Figure~\ref{fig:results} shows some visual results for 3 selected videos.
For comparison we show in Table~\ref{Tab:evaluationCD} also the results of two algorithms leading the CD ranking. The FTSG (Flux Tensor with Split Gaussian models) \cite{FTSG} algorithm is based on mixture of Gaussians which won the 2014 CD challenge. The PAWCS (Pixel-based Adaptive Word Consensus Segmenter) \cite{PAWCS} is a word-based approach to background modeling. While the raw results of DMD clearly cannot compete with the two mentioned highly optimized methods, it can be seen that simple post-processing can accelerate the performance substantially. Hence the object detection rate can be improved by learning from other background modeling methods and using for example, a more elaborate threshold, or integrating DMD into a system allowing background maintenance. 
\begin{table*}[ht]
	\centering
	\scalebox{0.6}{
		\begin{tabular}{ l l c c c c c c c c c c c} 
			\hline
			& \multicolumn{1}{l}{Measure}
			& \multicolumn{4}{c}{Baseline} & \multicolumn{2}{c}{Dynamic Background} & \multicolumn{2}{c}{Thermal}
			& \multicolumn{1}{c}{Average}
			\\
			\cmidrule(r){3-6}
			\cmidrule(r){7-8}
			\cmidrule(r){9-10}
			&  				& Highway & Pedestrians & PETS2006 & Office & Overpass & Canoe & Park & Lakeside  \\
			\hline
			\multirow{3}{*}{\rotatebox[origin=c]{0}{ \parbox{3.2cm}{ PAWCS \\ St-Charles et al. \cite{PAWCS} }     }} 
			& Recall 		& 0.952 & 0.961 & 0.945 & 0.905 & 0.961 & 0.947 & 0.899  & 0.520 & -
			\\
			& Precision    	& 0.935 & 0.931 & 0.919 & 0.972 & 0.957  & 0.929 & 0.768 & 0.752 & -
			\\
			& F-Measure 	& 0.944 & 0.946 & 0.932 & 0.937 & 0.959  & 0.938 & 0.829 & 0.615 & 0.887 
			\\ \hline
			
			\multirow{3}{*}{\rotatebox[origin=c]{0}{ \parbox{3cm}{ FTSG \\ Wang et al. \cite{wang2014static}} }} 
			& Recall 		& 0.956 & 0.979 & 0.963 & 0.908 & 0.944 & 0.913 & 0.666 & 0.228 & -
			\\
			& Precision    	& 0.934 & 0.890 & 0.883 & 0.961 & 0.941 & 0.985 & 0.724 & 0.960  &  -
			\\
			& F-Measure 	& 0.945 & 0.932 & 0.921 & 0.934 & 0.943 & 0.948 & 0.694 & 0.369 & 0.835 
			\\ \hline
			
			\multirow{3}{*}{\rotatebox[origin=c]{0}{ \parbox{3cm}{ rDMD \\ } }} 
			& Recall 		& 0.810 & 0.943 & 0.680 & 0.482 & 0.797 & 0.854 & 0.736  & 0.680 &  -
			\\
			& Precision    	& 0.789 & 0.756 & 0.703 & 0.560 & 0.194 & 0.201 & 0.610 & 0.448 &  -
			\\
			& F-Measure 	& 0.799 & 0.839 & 0.691 & 0.518 & 0.312 & 0.325 & 0.667 & 0.540 &  0.586
			\\ \hline
			
			\multirow{3}{*}{\rotatebox[origin=c]{0}{ \parbox{3cm}{rDMD \\ (with median filter) }  }} 
			& Recall 		& 0.901 & 0.976 & 0.681 & 0.551 & 0.778 & 0.900 & 0.816 & 0.655 &  -
			\\
			& Precision    	& 0.899 & 0.945 & 0.713 & 0.642 & 0.929 & 0.937 & 0.744 & 0.571 &  -
			\\
			& F-Measure 	& 0.900 & 0.960 & 0.696 & 0.593 & 0.847 & 0.918 & 0.779 & 0.610 &  0.788
			\\ \hline			
			
			\hline
		\end{tabular}
	}
	\caption{Evaluation results of eight real videos from the CD dataset. For comparison, the results of two algorithms from the CD ranking are presented.}
	\label{Tab:evaluationCD}
\end{table*}

\subsection{Computational performance}
We now evaluate the computational performance of rSVD and rDMD algorithm respectively. Our implementations are written in \textit{Python}\footnote{Python Software Foundation. Python Language Reference, version 2.7. Available at http://www.python.org} using the multi-thread MKL (Intel Math Kernel Library) accelerated linear algebra library \textit{LAPACK}. For the GPU implementation we are using \textit{NVIDIA CUDA} in combination with the linear algebra libraries \textit{cuBLAS} \cite{CUDA} and \textit{CULA} \cite{CULA}. To allow the comparison of rSVD with the fast LMSVD \cite{Liu12limitedmemory} algorithm we have used \textit{Matlab}. All the computations were performed on a standard gaming notebook (Intel Core i7-5500U 2.4GHz, 8GB DDR3 L memory and NVIDIA GeForce GTX 950M). It is important to note that in order to achieve any computational advantage with rSVD the target rank has to be $k < \frac{n}{1.5}$, otherwise truncated SVD would be faster. Another requirement is that the matrix fits into the fast memory.   
  
\subsubsection{Randomized SVD}
\begin{figure*}[htp]
	\centering
	\begin{subfigure}{0.85\textwidth}
		\DeclareGraphicsExtensions{.eps}
		\includegraphics[width=1\textwidth]{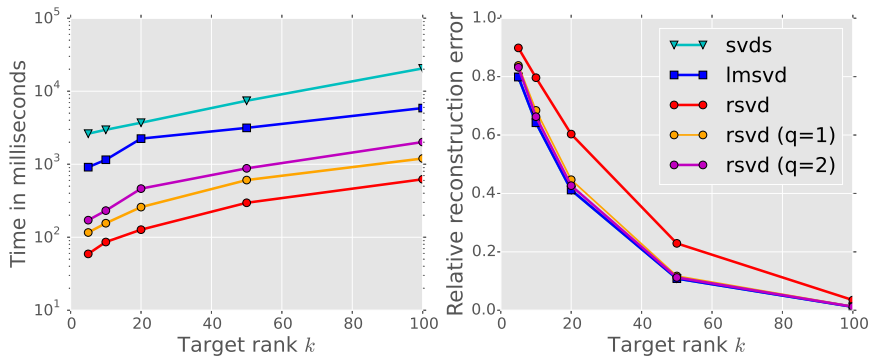}
		\caption{Square ($3000\times 3000$) random matrix.}
	\end{subfigure}

	\begin{subfigure}{0.85\textwidth}
		\DeclareGraphicsExtensions{.eps}
		\includegraphics[width=1\textwidth]{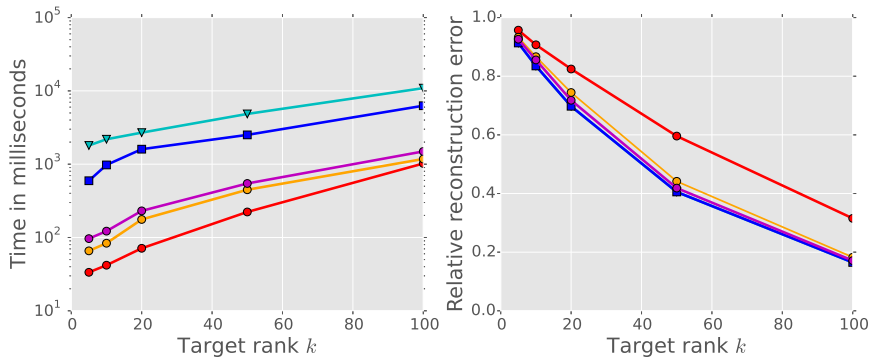}
		\caption{Thin ($1e5\times 500$) random matrix.}
	\end{subfigure}
	\caption{Computational performance of fast SVD algorithms on two random matrices for 5 different target ranks $k$. Randomized SVD outperforms partial SVD (svds) and LMSVD, while LMSVD is more accurate. Performing 1 or 2 subspace iterations improves the reconstruction error of rSVD considerably. The time is plotted on a log scale.}
	\label{fig:svdtiming}
\end{figure*}

Figure~\ref{fig:svdtiming} shows the computational time for rSVD with LMSVD and a partial SVD (svds) algorithm for two different sized matrices and a varying target rank. rSVD with one subspace iteration can achieve time savings of about a factor 10 to 30. Comparing to the LMSVD (with default options) the speed-up is about 5 to 8 times. However, the reconstruction error shows that the LMSVD algorithm is more precise, in particular when comparing to rSVD without subspace iterations.  

\subsubsection{Randomized DMD}
\begin{figure}[htp]
	\centering
	\DeclareGraphicsExtensions{.eps}
	\includegraphics[width=0.7\textwidth]{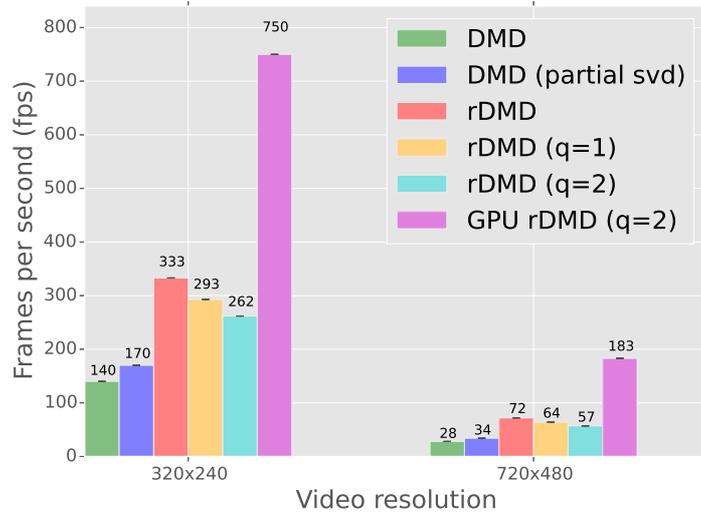}
	\caption{Comparison of the computational performance of DMD on real videos. Randomized DMD (even with subspace iterations) outperforms DMD using a truncated or partial SVD algorithm. In addition, using an GPU accelerated implementation can substantially increase the performance.}
	\label{fig:dmdtiming}
\end{figure}

For feasible real-time processing, our aim was to accelerate the ordinary DMD algorithm by using a randomized matrix algorithm for computing the SVD. Figure~\ref{fig:dmdtiming} shows the computational times we obtain with a randomized algorithm on videos with two different resolutions. The randomized version allows us to accelerate the computational time by about a factor of 2. Even more drastic is the acceleration using a GPU implementation. This allows processing of up to 180 frames per second for a 720x480 video. For a 320x240 video it can increase the frames per second from about 300 up to 750. The GPU accelerated DMD implementation benefits in particular from the fast computations of dot products, QR-decomposition and the fast generation of random numbers. Using a high-end graphic card can further improve the results, enabling real-time processing of HD 720 videos and beyond. The limitation of a GPU implementation is that the snapshot sequence has to fit into the graphic card's memory.

\section{Conclusion}\label{sec:conclusion}
We have presented a fast algorithm for computing the low-rank DMD using a randomized matrix algorithm. This subtle modification leads to substantial decreases in computational time, enabling DMD to process videos with high resolutions in real-time. Furthermore, the randomized version is of particular interest for parallel processing, e.g. GPU accelerated implementations. Randomized matrix algorithms may be beneficial for many other methods built on numerical linear algebra and in particular for those that use SVD. 

The suitability of DMD for motion detection has been evaluated on synthetic and real videos using statistical metrics. The results show that DMD can be seen as a fast approximation of RPCA. The results compared to other robust PCA algorithms are competitive, but the results of DMD compared to advanced statistical models leading the CD ranking are less optimal in terms of accuracy metrics. However, we have also shown that simple post-processing can enhance the results substantially. The performance can be further improved by adaptive thresholds or by integration into a background modeling system. This makes DMD interesting for applications where fast processing is more important then extremely high precision. 

In addition, we note that DMD is not a method purely designed for background modeling. DMD comes with a rich mathematical framework, which offers potential for interesting further research and applications beyond video. For example, interesting research directions are opened by multi-resolution dynamic mode decomposition \cite{mrdmd}, making the algorithm particularly interesting for the task of object tracking and motion estimation. Another direction is compressed DMD for fast background modeling. Both compressed and randomized DMD offer the opportunity to use importance sampling strategies to model a more robust background.

\section*{Acknowledgment}
For the many discussions and insightful remarks on dynamic mode decomposition we would like to thank J. Nathan Kutz and Steven L. Brunton. We would also like to express our gratitude to Leonid Sigal and the two anonymous reviewers for the many helpful comments to improve this paper.
N. Benjamin Erichson acknowledges support from the UK Engineering and Physical Sciences Research Council (EPSRC).

\bibliographystyle{elsarticle-num}
\bibliography{references}

\begin{thebibliography}{10}
\expandafter\ifx\csname url\endcsname\relax
  \def\url#1{\texttt{#1}}\fi
\expandafter\ifx\csname urlprefix\endcsname\relax\def\urlprefix{URL }\fi
\expandafter\ifx\csname href\endcsname\relax
  \def\href#1#2{#2} \def\path#1{#1}\fi

\bibitem{bouwmans2014robust}
T.~Bouwmans, E.~H. Zahzah, Robust pca via principal component pursuit: A review
  for a comparative evaluation in video surveillance, Computer Vision and Image
  Understanding 122 (2014) 22--34.
\newblock \href {http://dx.doi.org/10.1016/j.cviu.2013.11.009}
  {\path{doi:10.1016/j.cviu.2013.11.009}}.

\bibitem{Kutz2013}
J.~N. Kutz, Data-Driven Modeling and Scientific Computation: Methods for
  Complex Systems and Big Data, Oxford University Press, 2013.

\bibitem{Kutz2014dmd}
J.~Grosek, J.~N. Kutz, Dynamic mode decomposition for real-time
  background/foreground separation in video (2014).
\newblock \href {http://arxiv.org/abs/1404.7592} {\path{arXiv:1404.7592}}.

\bibitem{bouwmans2014traditional}
T.~Bouwmans, Traditional and recent approaches in background modeling for
  foreground detection: An overview, Computer Science Review 11-12 (2014)
  31--66.
\newblock \href {http://dx.doi.org/10.1016/j.cosrev.2014.04.001}
  {\path{doi:10.1016/j.cosrev.2014.04.001}}.

\bibitem{Sobralreview}
A.~Sobral, A.~Vacavant, A comprehensive review of background subtraction
  algorithms evaluated with synthetic and real videos, Computer Vision and
  Image Understanding 122 (2014) 4--21.
\newblock \href {http://dx.doi.org/10.1016/j.cviu.2013.12.005}
  {\path{doi:10.1016/j.cviu.2013.12.005}}.

\bibitem{oliver2000bayesian}
N.~Oliver, B.~Rosario, A.~Pentland, A bayesian computer vision system for
  modeling human interactions, Pattern Analysis and Machine Intelligence, IEEE
  Transactions on 22~(8) (2000) 831--843.
\newblock \href {http://dx.doi.org/10.1109/34.868684}
  {\path{doi:10.1109/34.868684}}.

\bibitem{bouwmans2009subspace}
T.~Bouwmans, Subspace learning for background modeling: A survey, Recent
  Patents on Computer Science 2~(3) (2009) 223--234.

\bibitem{bouwmans2015decomp}
T.~Bouwmans, A.~Sobral, S.~Javed, S.~K. Jung, E.-H. Zahzah, Decomposition into
  low-rank plus additive matrices for background/foreground separation: A
  review for a comparative evaluation with a large-scale dataset (2015).
\newblock \href {http://arxiv.org/abs/1511.01245} {\path{arXiv:1511.01245}}.

\bibitem{candes2011robust}
E.~J. Cand\`{e}s, X.~Li, Y.~Ma, J.~Wright, Robust principal component
  analysis?, Journal of the ACM 58~(3) (2011) 1--37.
\newblock \href {http://dx.doi.org/10.1145/1970392.1970395}
  {\path{doi:10.1145/1970392.1970395}}.

\bibitem{StableRPCA}
Z.~Zhou, X.~Li, J.~Wright, E.~Candes, Y.~Ma, Stable principal component
  pursuit, in: Information Theory Proceedings (ISIT), 2010 IEEE International
  Symposium on, 2010, pp. 1518--1522.
\newblock \href {http://dx.doi.org/10.1109/ISIT.2010.5513535}
  {\path{doi:10.1109/ISIT.2010.5513535}}.

\bibitem{Goldfarb}
D.~Goldfarb, S.~Ma, K.~Scheinberg, Fast alternating linearization methods for
  minimizing the sum of two convex functions, Mathematical Programming
  141~(1-2) (2013) 349--382.
\newblock \href {http://dx.doi.org/10.1007/s10107-012-0530-2}
  {\path{doi:10.1007/s10107-012-0530-2}}.

\bibitem{Becker}
S.~Becker, E.~J. Cand\`{e}s, M.~C. Grant, Templates for convex cone problems
  with applications to sparse signal recovery., Mathematical Programming
  Computation 3~(3) (2011) 165--218.
\newblock \href {http://dx.doi.org/10.1007/s12532-011-0029-5}
  {\path{doi:10.1007/s12532-011-0029-5}}.

\bibitem{guyon2012moving}
C.~Guyon, T.~Bouwmans, E.-H. Zahzah, Moving object detection by robust pca
  solved via a linearized symmetric alternating direction method, in: Advances
  in Visual Computing, Springer, 2012, pp. 427--436.

\bibitem{Liu12limitedmemory}
X.~Liu, Z.~Wen, Y.~Zhang, Limited memory block krylov subspace optimization for
  computing dominant singular value decompositions (2012).

\bibitem{LiuGauss}
X.~Liu, Z.~Wen, Y.~Zhang, An efficient gauss--newton algorithm for symmetric
  low-rank product matrix approximations, SIAM Journal on Optimization 25~(3)
  (2015) 1571--1608.
\newblock \href {http://dx.doi.org/10.1137/140971464}
  {\path{doi:10.1137/140971464}}.

\bibitem{frieze2004fast}
A.~Frieze, R.~Kannan, S.~Vempala, Fast monte-carlo algorithms for finding
  low-rank approximations, Journal of the ACM 51~(6) (2004) 1025--1041.

\bibitem{drineas2006fast}
P.~Drineas, R.~Kannan, M.~W. Mahoney, Fast monte carlo algorithms for matrices
  ii: Computing a low-rank approximation to a matrix, SIAM Journal on Computing
  36~(1) (2006) 158--183.

\bibitem{Martinsson201147}
P.-G. Martinsson, V.~Rokhlin, M.~Tygert, A randomized algorithm for the
  decomposition of matrices, Applied and Computational Harmonic Analysis 30~(1)
  (2011) 47--68.
\newblock \href {http://dx.doi.org/10.1016/j.acha.2010.02.003}
  {\path{doi:10.1016/j.acha.2010.02.003}}.

\bibitem{halko2011rand}
N.~Halko, P.~G. Martinsson, J.~A. Tropp, Finding structure with randomness:
  Probabilistic algorithms for constructing approximate matrix decompositions,
  SIAM Review 53~(2) (2011) 217--288.
\newblock \href {http://dx.doi.org/10.1137/090771806}
  {\path{doi:10.1137/090771806}}.

\bibitem{gu2015subspace}
M.~Gu, Subspace iteration randomization and singular value problems, SIAM
  Journal on Scientific Computing 37~(3) (2015) A1139--A1173.
\newblock \href {http://dx.doi.org/10.1137/130938700}
  {\path{doi:10.1137/130938700}}.

\bibitem{Mahoney2011}
M.~W. Mahoney, Randomized algorithms for matrices and data, Foundations and
  Trends in Machine Learning 3~(2) (2011) 123--224.
\newblock \href {http://dx.doi.org/10.1561/2200000035}
  {\path{doi:10.1561/2200000035}}.

\bibitem{Zhou11godec}
T.~Zhou, D.~Tao, Godec: Randomized low-rank \& sparse matrix decomposition in
  noisy case, in: International Conference on Machine Learning, ICML, 2011, pp.
  1--8.

\bibitem{watkins2004fundamentals}
D.~S. Watkins, Fundamentals of Matrix Computations, 2nd Edition, John Wiley \&
  Sons, Inc., New York, NY, USA, 2002.

\bibitem{voronin2015rsvdpack}
S.~Voronin, P.-G. Martinsson, Rsvdpack: Subroutines for computing partial
  singular value decompositions via randomized sampling on single core, multi
  core, and gpu architectures (2015).
\newblock \href {http://arxiv.org/abs/1502.05366} {\path{arXiv:1502.05366}}.

\bibitem{demmel1997applied}
J.~Demmel, Applied Numerical Linear Algebra, Society for Industrial and Applied
  Mathematics, 1997.
\newblock \href {http://dx.doi.org/10.1137/1.9781611971446}
  {\path{doi:10.1137/1.9781611971446}}.

\bibitem{woolfe2008fast}
F.~Woolfe, E.~Liberty, V.~Rokhlin, M.~Tygert, A fast randomized algorithm for
  the approximation of matrices, Applied and Computational Harmonic Analysis
  25~(3) (2008) 335--366.

\bibitem{schmid2010dynamic}
P.~J. Schmid, Dynamic mode decomposition of numerical and experimental data,
  Journal of Fluid Mechanics 656 (2010) 5--28.
\newblock \href {http://dx.doi.org/10.1017/S0022112010001217}
  {\path{doi:10.1017/S0022112010001217}}.

\bibitem{rowley2009spectral}
C.~W. Rowley, I.~Mezi\`{c}, S.~Bagheri, P.~Schlatter, D.~S. Hennigson, Spectral
  analysis of nonlinear flows, Journal of Fluid Mechanics 641 (2009) 115--127.
\newblock \href {http://dx.doi.org/10.1017/S0022112009992059}
  {\path{doi:10.1017/S0022112009992059}}.

\bibitem{kutz2015multi}
J.~N. Kutz, X.~Fu, S.~L. Brunton, Multi-resolution dynamic mode decomposition
  (2015).
\newblock \href {http://arxiv.org/abs/1506.00564} {\path{arXiv:1506.00564}}.

\bibitem{tu2013dynamic}
J.~H. {Tu}, C.~W. {Rowley}, D.~M. {Luchtenburg}, S.~L. {Brunton}, J.~N. {Kutz},
  On dynamic mode decomposition: Theory and applications (2013).
\newblock \href {http://arxiv.org/abs/1312.0041} {\path{arXiv:1312.0041}}.

\bibitem{golub1965calculating}
G.~Golub, W.~Kahan, Calculating the singular values and pseudo-inverse of a
  matrix, Journal of the Society for Industrial \& Applied Mathematics, Series
  B: Numerical Analysis 2~(2) (1965) 205--224.

\bibitem{jovanovic2012low}
M.~Jovanovic, P.~Schmid, J.~Nichols, Low-rank and sparse dynamic mode
  decomposition, Center for Turbulence Research Annual Research Briefs (2012)
  139--152.

\bibitem{vacavant2013benchmark}
A.~Vacavant, T.~Chateau, A.~Wilhelm, L.~Lequievre, A benchmark dataset for
  outdoor foreground/background extraction, in: Computer Vision--ACCV 2012
  Workshops, Springer, 2013, pp. 291--300.

\bibitem{wang2014cdnet}
Y.~Wang, P.-M. Jodoin, F.~Porikli, J.~Konrad, Y.~Benezeth, P.~Ishwar, Cdnet
  2014: an expanded change detection benchmark dataset, in: IEEE Workshop on
  Computer Vision and Pattern Recognition, IEEE, 2014, pp. 393--400.

\bibitem{benezeth2010comparative}
Y.~Benezeth, P.-M. Jodoin, B.~Emile, H.~Laurent, C.~Rosenberger, Comparative
  study of background subtraction algorithms, Journal of Electronic Imaging
  19~(3) (2010) --.
\newblock \href {http://dx.doi.org/10.1117/1.3456695}
  {\path{doi:10.1117/1.3456695}}.

\bibitem{fawcett2005ROC}
T.~Fawcett, An introduction to roc analysis, Pattern Recogn. Lett. 27~(8)
  (2006) 861--874.
\newblock \href {http://dx.doi.org/10.1016/j.patrec.2005.10.010}
  {\path{doi:10.1016/j.patrec.2005.10.010}}.

\bibitem{FTSG}
R.~Wang, F.~Bunyak, G.~Seetharaman, K.~Palaniappan, Static and moving object
  detection using flux tensor with split gaussian models, in: Computer Vision
  and Pattern Recognition Workshops (CVPRW), 2014 IEEE Conference on, IEEE,
  2014, pp. 420--424.

\bibitem{PAWCS}
P.-L. St-Charles, G.-A. Bilodeau, R.~Bergevin, A self-adjusting approach to
  change detection based on background word consensus, in: Applications of
  Computer Vision (WACV), 2015 IEEE Winter Conference on, IEEE, 2015, pp.
  990--997.

\bibitem{wang2014static}
R.~Wang, F.~Bunyak, G.~Seetharaman, K.~Palaniappan, Static and moving object
  detection using flux tensor with split gaussian models, in: Computer Vision
  and Pattern Recognition Workshops (CVPRW), 2014 IEEE Conference on, IEEE,
  2014, pp. 420--424.

\bibitem{CUDA}
J.~Nickolls, I.~Buck, M.~Garland, K.~Skadron, Scalable parallel programming
  with cuda, Queue 6~(2) (2008) 40--53.
\newblock \href {http://dx.doi.org/10.1145/1365490.1365500}
  {\path{doi:10.1145/1365490.1365500}}.

\bibitem{CULA}
J.~R. Humphrey, D.~K. Price, K.~E. Spagnoli, A.~L. Paolini, E.~J. Kelmelis,
  Cula: hybrid gpu accelerated linear algebra routines (2010).
\newblock \href {http://dx.doi.org/10.1117/12.850538}
  {\path{doi:10.1117/12.850538}}.

\bibitem{mrdmd}
J.~N. Kutz, X.~Fu, S.~L. Brunton, Multi-resolution analysis of dynamical
  systems using dynamic mode decomposition, in: Proceedings of the World
  Congress on Engineering, Vol.~1, WCE, 2015, pp. 1--4.

\end{thebibliography}

\end{document}